# Attitude Reconstruction from Inertial Measurements: QuatFIter and Its Comparison with RodFIter

Yuanxin Wu, *Senior Member, IEEE* and Gongmin Yan

*Abstract*— RodFIter is a promising method of attitude reconstruction from inertial measurements based on the functional iterative integration of Rodrigues vector. The Rodrigues vector is used to encode the attitude in place of the popular rotation vector because it has a polynomial-like rate equation and could be cast into theoretically sound and exact integration. This paper further applies the approach of RodFIter to the unity-norm quaternion for attitude reconstruction, named QuatFIter, and shows that it is identical to the previous Picard-type quaternion method. The Chebyshev polynomial approximation and truncation techniques from the RodFIter are exploited to speed up its implementation. Numerical results demonstrate that the QuatFIter is comparable in accuracy to the RodFIter, although its convergence rate is relatively slower with respect to the number of iterations. Notably, the QuatFIter has about two times better computational efficiency, thanks to the linear quaternion kinematic equation.

## I. INTRODUCTION

Attitude computation by integrating angular velocity is an important way to acquire the attitude, rotation or orientation [1, 2]. Attitude information is essential to many areas like unmanned vehicle navigation and control, virtual/augmented reality, satellite communication, robotics, and computer vision [3]. The modern-day attitude algorithm structure in the inertial navigation field, established in 1970s [4, 5], has relied on a simplified rotation vector differential equation for incremental attitude update [6, 7]. In parallel, a number of fields prefer the quaternion to deal with attitude computation as it is singularity-free and has a linear differential equation, e.g., robotics [8, 9], space applications [10] and computational mathematics [11-13] where the structure-preserving attributes of geometric integration is preferred. It might be argued that the common numerical methods such as the Runge Kutta can also be used to implement the attitude integration but they are not norm-preserving and thus inevitably lead to accuracy loss in attitude computation [14]. However, the above-mentioned methods have no way to drive the non-commutativity

This work was supported in part by National Natural Science Foundation of China (61422311, 61673263) and Joint Fund of China Ministry of Education (6141A02022309).

Authors' address: Yuanxin Wu, Shanghai Key Laboratory of Navigation and Location-based Services, School of Electronic Information and Electrical Engineering, Shanghai Jiao Tong University, Shanghai, China, 200240, E-mail: (yuanx_wu@hotmail.com); Gongmin Yan, School of Automation, Northwestern Polytechnical University, Xi'an, China, 710072, E-mail: (yangongmin@nwpu.edu.cn).

error to zero with practically finite sampling intervals. It has long been believed by navigation practitioners that the modern-day attitude algorithm is already good enough for applications [6, 15]. However, dynamic situations like the actual military projectiles with complex high-speed rotations and the cold-atom interference gyroscopes of ultra-high precision on the horizon demand for more accurate attitude algorithms.

Recently in the navigation community, independent works on high-accurate attitude algorithms have appeared or been under way [16-18]. They share the same spirit of trying to accurately solve the attitude kinematic equation based on the fitted angular velocity polynomial function. The main difference among these works is the chosen attitude parameterization. The quaternion is employed in [18], which first demonstrated in the public literature the practical potential of the Picard-type successive approximation method. The direction cosine matrix (DCM) could be used instead by the Taylor series expansion [19]. These methods can be traced back to the Russian seminal work in 1990s [20], where they were respectively classified as Type I and Type II methods. In specific, the Type II method used the rotation vector instead of DCM. The three-component attitude parameterization is minimal and does not need to satisfy the inherent constraints of the redundant-component parameterizations, such as the quaternion with the unit-norm constraint and the DCM with the orthogonal and +1-determinant constraints. In view of the finite-polynomial-like differential equation of the three-component Rodrigues vector, Wu [16] proposes the RodFIter method to reconstruct the attitude, which makes a natural use of the iterative function integration of the Rodrigues vector's kinematic equation. It highlights the capability of analytical attitude reconstruction over the whole update interval and provably converges to the true attitude if only the angular velocity is exact. Unfortunately, all high-accurate attitude algorithms face the problem of high computational burden, especially for real-time applications. Hence, to improve the computational efficiency, Wu [17] further comes up with a substantially fast version of RodFIter at little expense of accuracy by exploiting the excellent property of Chebyshev polynomial. It reformulates the original RodFIter in terms of the iterative computation of the Rodrigues vector's Chebyshev polynomial coefficients and exerts the Chebyshev polynomial truncation. In principle, the idea of RodFIter could be extended to various attitude parameters including the quaternion, but a question worthy of being investigated is whether the unit-norm constraint of quaternion affects the accuracy to be achieved. It is right the concern that has prevented one from using quaternion for attitude computation before Yan's work [18].

The contribution of this paper is that the technique of iterative functional integration and Chebyshev polynomial truncation is applied to the quaternion kinematic equation (named as the QuatFIter hereafter) for attitude reconstruction. It is shown that the QuatFIter has about two times better computational efficiency than the RodFIter, thanks to the linear quaternion kinematic equation. The remainder of the paper is organized as follows: Section II is devoted to developing the QuatFIter in the framework of iterative functional integration and analyzing its convergence property. The QuatFIter's relationship to the Picard-type quaternion method is discussed. Section III improves the computational efficiency of the QuatFIter by appropriate Chebyshev polynomial truncation, and then analyzes the error characteristics accordingly. Section IV evaluates the QuatFIter in terms of convergence speed, accuracy and computation efficiency as compared with the RodFIter as well as two common numerical integration methods by numerical simulations. A brief summary is given in the last section of the paper.

## II. QuatFIter for Attitude Reconstruction

Quaternion is an extension of complex number and constitutes a non-commutative group [21]. Attitude quaternion $\mathbf{q}$ can be represented as a four-dimensional column vector of unit magnitude, i.e., $\mathbf{q} = \begin{bmatrix} s & \mathbf{\eta}^T \end{bmatrix}^T$, where $s$ is the scalar part and $\mathbf{\eta}$ is the vector part. If these two parts are regarded as a scalar quaternion and a vector quaternion, respectively, then quaternion can be alternatively written as $\mathbf{q} = s + \mathbf{\eta}$.

The attitude quaternion kinematic equation is related to the angular velocity as

$$\dot{\mathbf{q}} = \frac{1}{2} \mathbf{q} \circ \mathbf{\omega}, \tag{1}$$

where $\mathbf{\omega}$ is the angular velocity vector quaternion with zero scalar part formed by the three-dimensional angular velocity vector. With some abuse of symbols, a vector quaternion is taken equally as a three-dimensional column vector throughout the paper. The operator $\circ$ means the multiplication of quaternions, defined as

$$\mathbf{q}_1 \circ \mathbf{q}_2 = [\mathbf{q}_1]^+ \begin{bmatrix} s_2 \\ \mathbf{\eta}_2 \end{bmatrix} = [\mathbf{q}_2]^- \begin{bmatrix} s_1 \\ \mathbf{\eta}_1 \end{bmatrix}. \tag{2}$$

The two quaternion multiplication matrices, $[\mathbf{q}]^+$ and $[\mathbf{q}]^-$, are respectively defined by [3]

$$[\mathbf{q}]^+ \triangleq \begin{bmatrix} s & -\mathbf{\eta}^T \\ \mathbf{\eta} & s\mathbf{I}_3 + \mathbf{\eta}\times \end{bmatrix}, \quad [\mathbf{q}]^- \triangleq \begin{bmatrix} s & -\mathbf{\eta}^T \\ \mathbf{\eta} & s\mathbf{I}_3 - \mathbf{\eta}\times \end{bmatrix}, \tag{3}$$

where $\mathbf{\eta}\times$ is the anti-symmetric matrix formed by the vector part $\mathbf{\eta}$ and $\mathbf{I}_3$ is the identical matrix of dimension three. Note the quaternion kinematic equation could be written as a form of linear system in quaternion

$$\dot{\mathbf{q}} = [\mathbf{\omega}/2]^- \mathbf{q} \triangleq \mathbf{W}\mathbf{q}. \tag{4}$$

For the situation that the direction of the angular velocity remains fixed over a time interval $[0 \ t]$, the attitude quaternion solution can be solved as $\mathbf{q}(t) = \exp\left\{\int_0^t \mathbf{W} d\tau\right\} \mathbf{q}(0)$ [2]. However, there is no explicit solution for a general rotation of interest in this paper.

*A. QuatFIter and Its Convergence Property*

Without the loss of generality, consider the incremental/relative attitude over the time interval $[0\ \ t]$ with the initial quaternion given by $\mathbf{q}(0) = [1\ \ 0\ \ 0\ \ 0]^T \triangleq \mathbf{1}$. Integrating (1) yields

$$\mathbf{q} = \mathbf{q}(0) + \frac{1}{2}\int_0^t \mathbf{q} \circ \boldsymbol{\omega}\, d\tau = \mathbf{1} + \frac{1}{2}\int_0^t \mathbf{q} \circ \boldsymbol{\omega}\, d\tau. \tag{5}$$

Applying the functional iteration integration technique [16], the quaternion can be iteratively computed as

$$\mathbf{q}_{l+1} = \mathbf{1} + \frac{1}{2}\int_0^t \mathbf{q}_l \circ \boldsymbol{\omega}\, d\tau, \quad l \geq 0. \tag{6}$$

with some chosen initial attitude quaternion function, say an identity quaternion, namely, $\mathbf{q}_0(t) \equiv [1\ \ 0\ \ 0\ \ 0]^T$. It can be readily checked that the above iteration process will produce such a sequence of quaternion functions as

$$\begin{aligned}
\mathbf{q}_0 &= \mathbf{1} \\
\mathbf{q}_1 &= \mathbf{1} + \frac{1}{2}\int_0^t \mathbf{q}_0 \circ \boldsymbol{\omega}\, d\tau = \mathbf{1} + \frac{1}{2}\int_0^t \boldsymbol{\omega}\, d\tau \\
\mathbf{q}_2 &= \mathbf{1} + \frac{1}{2}\int_0^t \mathbf{q}_1 \circ \boldsymbol{\omega}\, d\tau = \mathbf{1} + \frac{1}{2}\int_0^t \left(\mathbf{1} + \frac{1}{2}\int_0^{\tau_1} \boldsymbol{\omega}\, d\tau_2\right) \circ \boldsymbol{\omega}\, d\tau_1 = \mathbf{1} + \frac{1}{2}\int_0^t \boldsymbol{\omega}\, d\tau + \frac{1}{2^2}\int_0^t \left(\int_0^{\tau_1} \boldsymbol{\omega}\, d\tau_2\right) \circ \boldsymbol{\omega}\, d\tau_1 \\
&\vdots \\
\mathbf{q}_l &= \mathbf{1} + \frac{1}{2}\int_0^t \mathbf{q}_{l-1} \circ \boldsymbol{\omega}\, d\tau = \mathbf{1} + \frac{1}{2}\int_0^t \boldsymbol{\omega}\, d\tau + \cdots + \frac{1}{2^l}\int_0^t \left(\int_0^{\tau_1} \cdots \left(\int_0^{\tau_{l-1}} \boldsymbol{\omega}\, d\tau_l\right) \cdots \circ \boldsymbol{\omega}\, d\tau_2\right) \circ \boldsymbol{\omega}\, d\tau_1
\end{aligned} \tag{7}$$

For a general linear system of the form (4), there exists a general convergence result stating that the iterative process (7) converges on any infinite time interval [22]. In the sequel, we will adapt the main convergence analysis therein to the QuatFIter for the sake of being self-contained.

*Theorem 1*: Given the true angular velocity function $\boldsymbol{\omega}$ over the interval $[0\ \ t]$, the iterative process as given in (7) converges to the true attitude quaternion function solution to (1) for bounded $t\sup|\boldsymbol{\omega}|$.

Proof. For the convenience of convergence analysis, the sequence of quaternion functions is written as

$$\mathbf{q}_l = \mathbf{q}_0 + \sum_{j=0}^{l-1}(\mathbf{q}_{j+1} - \mathbf{q}_j), \quad l \geq 1 \tag{8}$$

Denote $\alpha \triangleq \sup_{\tau \in [0\ \ t]}|\boldsymbol{\omega}(\tau)|$, where $|\cdot|$ denotes the vector norm. Then,

$$|\mathbf{q}_0| = 1$$

$$|\mathbf{q}_1 - \mathbf{q}_0| = \left|\frac{1}{2}\int_0^t \boldsymbol{\omega}\, d\tau\right| = \left|\int_0^t \mathbf{W}[1\ 0\ 0\ 0]^T d\tau\right| \le \int_0^t \|\mathbf{W}\| d\tau = \int_0^t |\boldsymbol{\omega}| d\tau \le \alpha t$$

$$|\mathbf{q}_2 - \mathbf{q}_1| = \left|\frac{1}{2}\int_0^t (\mathbf{q}_1 - \mathbf{q}_0)\circ \boldsymbol{\omega}\, d\tau\right| = \left\|\int_0^t \mathbf{W}(\mathbf{q}_1 - \mathbf{q}_0) d\tau\right\| \le \int_0^t \|\mathbf{W}\| |\mathbf{q}_1 - \mathbf{q}_0| d\tau \le \int_0^t \alpha^2 \tau\, d\tau = \frac{1}{2}\alpha^2 t^2 \quad (9)$$

$$\vdots$$

$$|\mathbf{q}_{j+1} - \mathbf{q}_j| = \left|\frac{1}{2}\int_0^t (\mathbf{q}_j - \mathbf{q}_{j-1})\circ \boldsymbol{\omega}\, d\tau\right| \le \frac{1}{(j+1)!}(\alpha t)^{j+1}$$

where $\|\cdot\|$ denotes the matrix Frobenius norm, namely the root of the sum of all matrix elements squared. Note that the sum of the bounds on the right side $1 + \alpha t + \frac{1}{2}\alpha^2 t^2 + \cdots + \frac{1}{(j+1)!}(\alpha t)^{j+1} + \cdots = e^{\alpha t}$. According to the Weierstrass M-Test (see Appendix), the iterative process (7) converges uniformly and absolutely on the interval $[0\ t]$. Finally, it can be readily checked that the limit of the quaternion function sequence satisfies (1), i.e., $\frac{d\mathbf{q}_\infty}{dt} = \frac{1}{2}\mathbf{q}_\infty \circ \boldsymbol{\omega}$. It means that $\mathbf{q}_\infty$ is the solution to the quaternion kinematic equation (1), that it to say, the iterative process (7) converges to the true attitude quaternion function.

∎

For those fixed-axis rotation cases that the **W**-commutativity condition holds, the limit of the quaternion function sequence is reduced to $\mathbf{q}_\infty = \exp\left\{\int_0^t \mathbf{W}\, d\tau\right\}\mathbf{q}(0)$ (see *Proposition 1* in Appendix). Recall that for the RodFIter based on the Rodrigues vector, the sufficient convergence condition we have proved is $t\sup|\boldsymbol{\omega}| < 2$ [16], while the QuatFIter converges for any bounded $t\sup|\boldsymbol{\omega}|$. This advantage of global convergence region comes as a result of the linear kinematic equation of the attitude quaternion. It may be argued that the region $t\sup|\boldsymbol{\omega}| < 2$ is already sufficient for practical applications using the RodFIter for incremental attitude update, because the Rodrigues vector is by definition singular at rotation angles $\pm\pi$ [3].

*B. Relation to Picard-type Successive Approximation Method*

The Picard-type successive approximation method [18, 20] repeatedly uses (5) to derive the approximation of the attitude quaternion. In specific, it substitutes (5) repeatedly into the right side of the resultant

$$\begin{aligned}\mathbf{q} &= \mathbf{1} + \frac{1}{2}\int_0^t \mathbf{q}\circ\boldsymbol{\omega}\, d\tau \\ &= \mathbf{1} + \frac{1}{2}\int_0^t \left(\mathbf{1} + \frac{1}{2}\int_0^{\tau_1}\mathbf{q}\circ\boldsymbol{\omega}\, d\tau_2\right)\circ\boldsymbol{\omega}\, d\tau_1 \\ &= \mathbf{1} + \frac{1}{2}\int_0^t \boldsymbol{\omega}\, d\tau + \frac{1}{2^2}\int_0^t\left(\int_0^{\tau_1}\mathbf{q}\circ\boldsymbol{\omega}\, d\tau_2\right)\circ\boldsymbol{\omega}\, d\tau_1\end{aligned} \quad (10)$$

A second substitution of (5) into the above gives

$$\mathbf{q} = \mathbf{1} + \frac{1}{2}\int_0^t \boldsymbol{\omega}\, d\tau + \frac{1}{2^2}\int_0^t \left( \int_0^{\tau_1} \left( \mathbf{1} + \frac{1}{2}\int_0^{\tau_2} \mathbf{q} \circ \boldsymbol{\omega}\, d\tau_3 \right) \circ \boldsymbol{\omega}\, d\tau_2 \right) \circ \boldsymbol{\omega}\, d\tau_1$$
$$= \mathbf{1} + \frac{1}{2}\int_0^t \boldsymbol{\omega}\, d\tau + \frac{1}{2^2}\int_0^t \left( \int_0^{\tau_1} \boldsymbol{\omega}\, d\tau_2 \right) \circ \boldsymbol{\omega}\, d\tau_1 + \frac{1}{2^3}\int_0^t \left( \int_0^{\tau_1} \left( \int_0^{\tau_2} \mathbf{q} \circ \boldsymbol{\omega}\, d\tau_3 \right) \circ \boldsymbol{\omega}\, d\tau_2 \right) \circ \boldsymbol{\omega}\, d\tau_1$$
(11)

Of course, we might substitute (10) instead of (5) into (10) and it yields

$$\mathbf{q} = \mathbf{1} + \frac{1}{2}\int_0^t \boldsymbol{\omega}\, d\tau + \frac{1}{2^2}\int_0^t \left( \int_0^{\tau_1} \left( \mathbf{1} + \frac{1}{2}\int_0^{\tau_2} \boldsymbol{\omega}\, dt + \frac{1}{2^2}\int_0^{\tau_2} \left( \int_0^{\tau_3} \mathbf{q} \circ \boldsymbol{\omega}\, d\tau_4 \right) \circ \boldsymbol{\omega}\, d\tau_3 \right) \circ \boldsymbol{\omega}\, d\tau_2 \right) \circ \boldsymbol{\omega}\, d\tau_1$$
$$= \mathbf{1} + \frac{1}{2}\int_0^t \boldsymbol{\omega}\, d\tau + \frac{1}{2^2}\int_0^t \left( \int_0^{\tau_1} \boldsymbol{\omega}\, d\tau_2 \right) \circ \boldsymbol{\omega}\, d\tau_1$$
$$+ \frac{1}{2^3}\int_0^t \left( \int_0^{\tau_1} \left( \int_0^{\tau_2} \boldsymbol{\omega}\, d\tau_3 \right) \circ \boldsymbol{\omega}\, d\tau_2 \right) \circ \boldsymbol{\omega}\, d\tau_1 + \frac{1}{2^4}\int_0^t \left( \int_0^{\tau_1} \left( \int_0^{\tau_2} \left( \int_0^{\tau_3} \mathbf{q} \circ \boldsymbol{\omega}\, d\tau_4 \right) \circ \boldsymbol{\omega}\, d\tau_3 \right) \circ \boldsymbol{\omega}\, d\tau_2 \right) \circ \boldsymbol{\omega}\, d\tau_1$$
(12)

Anyway, the Picard-type method [18, 20] takes the sum of all terms but the last one as an approximation of the attitude quaternion, which is identical to the QuatFIter sequence (7). Therefore, the last residual term containing the unknown attitude quaternion explicitly quantifies the approximation error of the QuatFIter sequence (7).

Denote the additive error of the quaternion function sequence (7) from the true quaternion function as $\delta \mathbf{q}_l \triangleq \mathbf{q}_l - \mathbf{q}$. Using (12), it can be derived that

$$|\delta \mathbf{q}_l| = \left| \int_0^t \left( \int_0^{\tau_1} \cdots \left( \int_0^{\tau_{l-1}} \mathbf{q} \circ \frac{\boldsymbol{\omega}}{2}\, d\tau_l \right) \cdots \circ \frac{\boldsymbol{\omega}}{2}\, d\tau_2 \right) \circ \frac{\boldsymbol{\omega}}{2}\, d\tau_1 \right|$$
$$= \left| \int_0^t \int_0^{\tau_1} \cdots \int_0^{\tau_{l-1}} \mathbf{W}(\tau_1)\mathbf{W}(\tau_2)\cdots\mathbf{W}(\tau_l)\mathbf{q}\, d\tau_l \cdots d\tau_2 d\tau_1 \right|$$
$$\leq \int_0^t \int_0^{\tau_1} \cdots \int_0^{\tau_{l-1}} \|\mathbf{W}(\tau_1)\|\|\mathbf{W}(\tau_2)\|\cdots\|\mathbf{W}(\tau_l)\||\mathbf{q}|\, d\tau_l \cdots d\tau_2 d\tau_1$$
$$\leq \int_0^t \int_0^{\tau_1} \cdots \int_0^{\tau_{l-1}} (\sup|\boldsymbol{\omega}|)^l\, d\tau_l \cdots d\tau_2 d\tau_1 = \frac{(t\sup|\boldsymbol{\omega}|)^l}{l!}$$
(13)

which approaches zero as $l \to \infty$ for any bounded $t\sup|\boldsymbol{\omega}|$. It can be regarded as an alternative proof of *Theorem 1*.

III. FAST QUATFITER BY CHEBYSHEV POLYNOMIAL APPROXIMATION AND TRUNCATION

*A. Angular Velocity Fitting by Chebyshev Polynomial*

The Chebyshev polynomial of the first kind is defined over the interval $[-1\ \ 1]$ by the recurrence relation as

$$F_0(x) = 1,\ F_1(x) = x,\ F_{i+1}(x) = 2xF_i(x) - F_{i-1}(x),$$
(14)

where $F_i(x)$ is the $i^{\text{th}}$-degree Chebyshev polynomial of the first kind. For any $j, k \geq 0$, the Chebyshev polynomial of first kind satisfies the equality [23] as follows:

$$F_j(\tau)F_k(\tau) = \frac{1}{2}\left(F_{j+k}(\tau) + F_{|j-k|}(\tau)\right).$$
(15)

Then the integrated $i^{th}$-degree Chebyshev polynomial can be expressed as a linear combination of Chebyshev polynomials, given by

$$G_{i,[-1\ \tau]} = \int_{-1}^{\tau} F_i(\tau) d\tau = \begin{cases} \left(\dfrac{F_{i+1}(\tau)}{2(i+1)} - \dfrac{F_{|i-1|}(\tau)}{2(i-1)}\right) - \dfrac{(-1)^i}{i^2-1} F_0(\tau) & \text{for } i \neq 1, \\ \dfrac{F_{i+1}(\tau)}{4} - \dfrac{F_0(\tau)}{4} & \text{for } i = 1, \end{cases} \quad (16)$$

where $F_i(-1) = (-1)^i$.

Assume discrete angular velocity or angular increment measurements at time instants $t_k$ ($k = 1, 2, \ldots N$) by a triad of gyroscopes. In order to apply the Chebyshev polynomials, the actual time interval is mapped onto $[-1\ 1]$ by letting $t = \dfrac{t_N}{2}(1+\tau)$. The RodFIter [16, 17] employs the Chebyshev polynomials to fit the angular velocity polynomial as

$$\hat{\boldsymbol{\omega}} = \sum_{i=0}^{n} \mathbf{c}_i F_i(\tau), \quad n \leq N-1. \quad (17)$$

where the coefficient $\mathbf{c}_i$ is determined by the least-square method using the discrete angular velocity or angular increment measurements. With the converging condition of *Theorem 1* in mind, substituting (17) into the iteration process (6) is supposed to well reconstruct the attitude quaternion function as a finite polynomial, as the group of polynomials are closed under elementary arithmetic operations.

*B. Fast QuatFIter by Iterative Computation of Chebyshev Polynomial Coefficients*

Next we will reformulate the QuatFIter as the iterative computation of Chebyshev polynomial coefficients and perform the analysis of computational complexity. Fast QuatFIter implementation will be achieved by appropriate truncation of Chebyshev polynomials. If the angular velocity is smooth, the absolute value of Chebyshev coefficients $\mathbf{c}_i$ in (17) will decrease exponentially [23]. Assume the quaternion estimate at the *l*-th iteration is given by a weighted sum of Chebyshev polynomials, say

$$\mathbf{q}_l \triangleq \sum_{i=0}^{m_l} \mathbf{b}_{l,i} F_i(\tau), \quad (18)$$

where $m_l$ is the maximum degree and $\mathbf{b}_{l,i}$ is the coefficient of $i^{th}$-degree Chebyshev polynomial at the *l*-th iteration. The integral in (6) is transformed to that over the mapped interval of Chebyshev polynomials, that is,

$$\mathbf{q}_{l+1} = \mathbf{1} + \frac{1}{2}\int_0^t \mathbf{q}_l \circ \boldsymbol{\omega}\, dt = \mathbf{1} + \frac{t_N}{4}\int_{-1}^{\tau} \mathbf{q}_l \circ \boldsymbol{\omega}\, d\tau. \quad (19)$$

Substituting (17)-(19), the integral on the right side has the form

$$\int_{-1}^{\tau} \mathbf{q}_l \circ \boldsymbol{\omega} d\tau = \int_{-1}^{\tau} \left[ \sum_{i=0}^{m_l} \mathbf{b}_{l,i} F_i(\tau) \right] \circ \sum_{j=0}^{n} \mathbf{c}_j F_j(\tau) d\tau = \sum_{i=0}^{m_l} \sum_{j=0}^{n} \mathbf{b}_{l,i} \circ \mathbf{c}_j \int_{-1}^{\tau} F_i(\tau) F_j(\tau) d\tau$$
$$= \frac{1}{2} \sum_{i=0}^{m_l} \sum_{j=0}^{n} \mathbf{b}_{l,i} \circ \mathbf{c}_j \int_{-1}^{\tau} \left( F_{i+j}(\tau) + F_{|i-j|}(\tau) \right) d\tau = \frac{1}{2} \sum_{i=0}^{m_l} \sum_{j=0}^{n} \mathbf{b}_{l,i} \circ \mathbf{c}_j \left( G_{i+j,[-1\ \tau]} + G_{|i-j|,[-1\ \tau]} \right).$$
(20)

So (19) is expressed as

$$\mathbf{q}_{l+1} = \mathbf{1} + \frac{t_N}{8} \sum_{i=0}^{m_l} \sum_{j=0}^{n} \mathbf{b}_{l,i} \circ \mathbf{c}_j \left( G_{i+j,[-1\ \tau]} + G_{|i-j|,[-1\ \tau]} \right) \triangleq \sum_{i=0}^{m_{l+1}} \mathbf{b}_{l+1,i} F_i(\tau),$$
(21)

where $m_{l+1} = m_l + n + 1$ with $m_0 = 0$. It can be readily verified that $m_l = (n+1)l$. As a quaternion multiplication (2) involves 16 scalar multiplications, the computational complexity of (21) in terms of scalar multiplications is proportional to $\mathrm{O}(16nm_l) = \mathrm{O}(16n(n+1)l)$. For instance, for $n=8$ at the 7th iteration, $m_l = 63$ and the QuatFIter's computational complexity is proportional to $\mathrm{O}(8000)$.

The computational burden can be reduced by means of appropriate polynomial truncation. Regarding (17), the angular velocity is fitted by a Chebyshev polynomial of degree $n$ in time and thus the first integral of (7), namely, the integrated angular velocity, is accurate up to degree $n+1$. It is reasonable to abandon those polynomials of higher order than some prescribed threshold, say $n_T$, to save computation

$$\hat{\mathbf{q}}_l = \sum_{i=0}^{n_T} \mathbf{b}_{l,i} F_i(\tau).$$
(22)

According to the coefficient-decreasing and ±1-bounded properties of Chebyshev polynomials [23], the truncation error is bounded by the coefficient of the first neglected Chebyshev polynomial, namely, $|\delta \mathbf{q}_l^t| \leq |\mathbf{b}_{l,n_T+1}|$. The superscript '$t$' denotes that the error is owed to the polynomial truncation.

The iteration (21) becomes

$$\hat{\mathbf{q}}_{l+1} = \mathbf{1} + \frac{t_N}{8} \sum_{i=0}^{n_T} \sum_{j=0}^{n} \mathbf{b}_{l,i} \circ \mathbf{c}_j \left( G_{i+j,[-1\ \tau]} + G_{|i-j|,[-1\ \tau]} \right) \triangleq \sum_{i=0}^{n_T+n+1} \mathbf{b}_{l+1,i} F_i(\tau) \overset{\text{polynomial truncation}}{\approx} \sum_{i=0}^{n_T} \mathbf{b}_{l+1,i} F_i(\tau),$$
(23)

where the last approximation is due to the truncation at each iteration. By way of polynomial truncation, the computational complexity is further reduced to $\mathrm{O}(16nn_T)$. For instance, it will be proportional to $\mathrm{O}(1200)$ for $n_T = n+1$ and $n=8$ at each iteration. In contrast, the fast RodFIter's computational complexity is $\mathrm{O}(6nn_T + 6nn_T^2) = \mathrm{O}(4300)$ at each iteration [17]. The significant reduction of 3~4 times is owed to the quaternion's linear rate relation to the quaternion itself and the angular velocity as shown in (1), while the Rodrigues vector rate equation is quadratic with respect to the Rodrigues vector itself [16, 17]. Table I lists the main steps of the QuatFIter and compares with the RodFIter for easy reference. It should be noted that the

Table I. QuatFIter and RodFIter

| | **QuatFIter** | **RodFIter** |
|---|---|---|
| Step 1: | *Chebyshev polynomial-based angular velocity fitting with N gyroscope measurements as (17)* $$\hat{\boldsymbol{\omega}} = \sum_{i=0}^{n} \mathbf{c}_i F_i(\tau), \ n \leq N-1$$ | |
| Step 2: | *Set* $l=0$ *and initial quaternion* $$\mathbf{q}_0 = \begin{bmatrix} 1 & 0 & 0 & 0 \end{bmatrix}^T$$ | *Set* $l=0$ *and initial Rodrigues vector* $$\mathbf{g}_0 = \begin{bmatrix} 0 & 0 & 0 \end{bmatrix}^T$$ |
| Step 3: | *Iteratively compute Chebyshev polynomial coefficients of attitude quaternion* $\hat{\mathbf{q}}_l$ *as (23) and truncate polynomial coefficients using prescribed order* $n_T$ $$\hat{\mathbf{q}}_{l+1} = \mathbf{1} + \frac{t_N}{8}\sum_{i=0}^{n_T}\sum_{j=0}^{n} \mathbf{b}_{l,i} \circ \mathbf{c}_j \left( G_{i+j,[-1\ \tau]} + G_{|i-j|,[-1\ \tau]} \right)$$ $$\triangleq \sum_{i=0}^{n_T+n+1} \mathbf{b}_{l+1,i} F_i(\tau) \overset{\text{polynomial truncation}}{\approx} \sum_{i=0}^{n_T} \mathbf{b}_{l+1,i} F_i(\tau)$$ | *Iteratively compute Chebyshev polynomial coefficients of Rodrigues vector* $\hat{\mathbf{g}}_l$ *and truncate polynomial coefficients using prescribed order* $n_T$ $$\hat{\mathbf{g}}_{l+1} = \frac{t_N}{2}\left( \begin{array}{c} \sum_{i=0}^{n}\mathbf{c}_i G_{i,[-1\ \tau]} + \frac{1}{4}\sum_{i=0}^{n_T}\sum_{j=0}^{n}\mathbf{b}_{l,i}\times\mathbf{c}_j\left( G_{i+j,[-1\ \tau]} + G_{|i-j|,[-1\ \tau]} \right) \\ +\frac{1}{16}\sum_{i=0}^{n_T}\sum_{j=0}^{n_T}\sum_{k=0}^{n}\mathbf{b}_{l,i}\mathbf{b}_{l,j}^T\mathbf{c}_k\begin{pmatrix}G_{i+j+k,[-1\ \tau]} + G_{|i+j-k|,[-1\ \tau]} \\ +G_{|i-j|+k,[-1\ \tau]} + G_{||i-j|-k|,[-1\ \tau]}\end{pmatrix} \end{array} \right)$$ $$\triangleq \sum_{i=0}^{2n_T+n+1} \mathbf{b}_{l+1,i} F_i(\tau) \overset{\text{polynomial truncation}}{\approx} \sum_{i=0}^{n_T} \mathbf{b}_{l+1,i} F_i(\tau)$$ |
| Step 4: | $l \leftarrow l+1$ *and go back to Step 3 until the maximum iteration or prescribed stop criterion* * *is reached* | |
| Step 5: | *Attitude quaternion obtained by normalization* $\hat{\mathbf{q}} = \hat{\mathbf{q}}_l / |\hat{\mathbf{q}}_l|$ | *Attitude quaternion* $\hat{\mathbf{q}} = \dfrac{2 + \hat{\mathbf{g}}_l}{\sqrt{4 + |\hat{\mathbf{g}}_l|^2}}$ |

*: The stop criterion could be set that the root mean square of the discrepancy of the polynomial coefficients between successive iterations is less than some threshold.

quaternion normalization is performed outside of the essential iteration (Steps 3-4), otherwise the quaternion estimate $\hat{\mathbf{q}}_l$ is not a finite polynomial any more. The RotFIter in [16] based on the approximate rotation vector kinematics also fits into the framework of iterative function integration but is not listed above, because it compares unfavorably with RodFIter therein.

*Theorem 2*: Given the true angular velocity function $\boldsymbol{\omega}$ over the interval $\begin{bmatrix} 0 & t \end{bmatrix}$, the fast QuatFIter (23) converges to the true attitude quaternion function up to the truncated polynomial degree $n_T$.

Proof. Interested readers are referred to Theorem 2 in [17] for some details.

## C. Error Analysis

It is obvious that the fast QuatFIter has three error sources: the angular velocity's error $\delta\boldsymbol{\omega}$, the initial iteration error $\delta\hat{\mathbf{q}}_0$, and the truncation error at the *l*-th iteration $\delta\mathbf{q}_l^t$. The additive quaternion error (to first order) of the fast QuatFIter propagates as

$$\begin{aligned}
|\delta\hat{\mathbf{q}}_1| &= \left| \frac{1}{2}\int_0^t \delta\hat{\mathbf{q}}_0 \circ \boldsymbol{\omega}\, dt + \frac{1}{2}\int_0^t \hat{\mathbf{q}}_0 \circ \delta\boldsymbol{\omega}\, dt + \delta\mathbf{q}_1^t \right| \\
&\leq \int_0^t \|\mathbf{W}\| |\delta\hat{\mathbf{q}}_0|\, dt + \frac{1}{2}\int_0^t \|[\overline{\delta\boldsymbol{\omega}}]\| |\hat{\mathbf{q}}_0|\, dt + |\delta\mathbf{q}_1^t| \\
&\leq t\sup|\boldsymbol{\omega}|\sup|\delta\hat{\mathbf{q}}_0| + t\sup|\delta\boldsymbol{\omega}|\sup|\hat{\mathbf{q}}_0| + |\delta\mathbf{q}_1^t|,
\end{aligned} \quad (24)$$

and

$$\begin{aligned}|\delta\hat{\mathbf{q}}_2| &\leq \int_0^t \|\mathbf{W}\||\delta\hat{\mathbf{q}}_1|dt + \frac{1}{2}\int_0^t \|[\bar{\delta\boldsymbol{\omega}}]\||\hat{\mathbf{q}}_1|dt + |\delta\mathbf{q}_2^t| \\
&\leq \int_0^t \sup|\boldsymbol{\omega}|\left(t\sup|\boldsymbol{\omega}|\sup|\delta\hat{\mathbf{q}}_0| + t\sup|\delta\boldsymbol{\omega}|\sup|\hat{\mathbf{q}}_0| + |\delta\mathbf{q}_1^t|\right)dt + t\sup|\delta\boldsymbol{\omega}|\sup|\hat{\mathbf{q}}_1| + |\delta\mathbf{q}_2^t| \\
&= \frac{t^2}{2}\left(\sup|\boldsymbol{\omega}|\right)^2 \sup|\delta\hat{\mathbf{q}}_0| + \frac{t^2}{2}\sup|\boldsymbol{\omega}|\sup|\delta\boldsymbol{\omega}|\sup|\hat{\mathbf{q}}_0| + t\sup|\delta\boldsymbol{\omega}|\sup|\hat{\mathbf{q}}_1| + t\sup|\boldsymbol{\omega}|\sup|\delta\mathbf{q}_1^t| + |\delta\mathbf{q}_2^t| \\
&\vdots \\
|\delta\hat{\mathbf{q}}_l| &\leq \int_0^t \|\mathbf{W}\||\delta\hat{\mathbf{q}}_{l-1}|dt + \frac{1}{2}\int_0^t \|[\bar{\delta\boldsymbol{\omega}}]\||\hat{\mathbf{q}}_{l-1}|dt + |\delta\mathbf{q}_l^t| \\
&\leq \frac{(t\sup|\boldsymbol{\omega}|)^l}{l!}\sup|\delta\hat{\mathbf{q}}_0| + t\sup|\delta\boldsymbol{\omega}|\sum_{k=0}^{l-1}\frac{(t\sup|\boldsymbol{\omega}|)^k}{(k+1)!}\sup|\hat{\mathbf{q}}_{l-1-k}| + \sum_{k=0}^{l-1}\frac{(t\sup|\boldsymbol{\omega}|)^k}{k!}\sup|\delta\hat{\mathbf{q}}_{l-k}^t|\end{aligned} \quad (25)$$

Among the above error components, the first term is owed to the initial quaternion error and quickly vanishes for large iterations for any bounded $t\sup|\boldsymbol{\omega}|$. It means the QuatFIter converges regardless of the initial quaternion function. The third term is owed to the polynomial truncation at each iteration, in which the weights of the early iterations are much smaller than those of later iterations, and thus can be approximated by the last truncation error, i.e.,

$$\sum_{k=0}^{l-1}\frac{(t\sup|\boldsymbol{\omega}|)^k}{k!}\sup|\delta\hat{\mathbf{q}}_{l-k}^t| \approx \sum_{k=0}^{l-1}\frac{(t\sup|\boldsymbol{\omega}|)^k}{k!}|\mathbf{b}_{l-k,n_T+1}| \approx |\mathbf{b}_{l,n_T+1}|. \quad (26)$$

The second term depends on the angular velocity error and the norm of the quaternion estimate, and it can be approximated by

$$t\sup|\delta\boldsymbol{\omega}|\sum_{k=0}^{l-1}\frac{(t\sup|\boldsymbol{\omega}|)^k}{(k+1)!}\sup|\hat{\mathbf{q}}_{l-1-k}| \approx t\sup|\delta\boldsymbol{\omega}|\sum_{k=0}^{l-1}\frac{(t\sup|\boldsymbol{\omega}|)^k}{(k+1)!} \approx t\sup|\delta\boldsymbol{\omega}|. \quad (27)$$

as $\sup|\hat{\mathbf{q}}| \approx 1$. Therefore, for a large number of iterations, the quaternion error of the fast QuatFIter in (23) is approximately bounded by

$$\sup|\delta\hat{\mathbf{q}}_l| \approx t\sup|\delta\boldsymbol{\omega}| + |\mathbf{b}_{l,n_T+1}|. \quad (28)$$

This indicates that the fast QuatFIter's error is generally dominated by the angular velocity error and the last truncation error as well. An ideal case is when $\delta\boldsymbol{\omega} = 0$, namely the angular velocity fitting error is zero, for which higher order of truncation means higher accuracy.

## IV. SIMULATION RESULTS

In this section, simulations are performed in the context of coning motion and constant rotation scenarios to evaluate the QuatFIter as compared with the RodFIter [16, 17].

The coning motion has explicit analytical expressions in angular velocity and the associated Rodrigues vector and attitude quaternion, so it is widely employed as a standard criterion for algorithm accuracy assessment in the inertial navigation field [1].

The angular velocity of the coning motion is given by $\boldsymbol{\omega} = \Omega\begin{bmatrix} -2\sin^2(\alpha/2) & -\sin(\alpha)\sin(\Omega t) & \sin(\alpha)\cos(\Omega t) \end{bmatrix}^T$. The corresponding true attitude quaternion is $\mathbf{q} = \dfrac{2+\mathbf{g}}{\sqrt{4+|\mathbf{g}|^2}}$, where the Rodrigues vector $\mathbf{g} = 2\tan(\alpha/2)\begin{bmatrix} 0 & \cos(\Omega t) & \sin(\Omega t) \end{bmatrix}^T$.

The coning angle is set to $\alpha = 10\,\mathrm{deg}$, the coning frequency $\Omega = 0.74\pi$ rad/s. The angular increment measurement from gyroscopes is assumed and the discrete sampling rate is nominally set to 100 Hz.

The following principal angle metric is used to quantify the attitude computation error

$$\varepsilon_{att} = 2\left|\left[\mathbf{q}^* \circ \hat{\mathbf{q}}\right]_{2:4}\right|, \tag{29}$$

where $\hat{\mathbf{q}}$ denotes the quaternion estimate, and the superscript * means the quaternion conjugate and the operator $[\cdot]_{2:4}$ takes the vector part of the error quaternion. Hereafter the order of the fitted angular velocity is uniformly set to $n = N-1$.

Figure 1 plots attitude errors of QuatFIter and RodFIter at each iteration in the first update interval for $N = 5$ (subfigures in the upper row) and $N = 8$ (subfigures in the lower row). Truncation order $n_T = n + 2$ is used for both, in contrast to $n_T = n + 1$ in [17], for the sake of marginally better accuracy. The quaternion estimate of QuatFIter is normalized right before using (29) to compute the principal angle error, as explicitly given by Step 5 in Table I. Lines with same colors indicate results of the same iteration. As for RodFIter denoted by decoration lines, the attitude errors reduce uniformly along with the iterations, while the QuatFIter's errors have a non-uniform error-reducing pace. For the $l$-th iteration, the QuatFIter's errors stay consistently above those of RodFIter; yet the QuatFIter's errors at the ($l$+1)-th iteration stay consistently below those of RodFIter at the $l$-th iteration. It means that RodFIter has a faster convergence rate with respect to the iterations, although they have almost identical final errors up to machine precision. The subfigures in the right column present the quaternion norm error from unity. It shows that the norm of the quaternion estimate approaches unity as the iteration goes. No clear relationship between the attitude error and the norm error can be identified. For instance, the quaternion norm errors at the 2$^{nd}$ and 3$^{rd}$ iterations are relatively close, but the attitude error at the 3$^{rd}$ iteration is reduced by about four orders from that at the 2$^{nd}$ iteration.

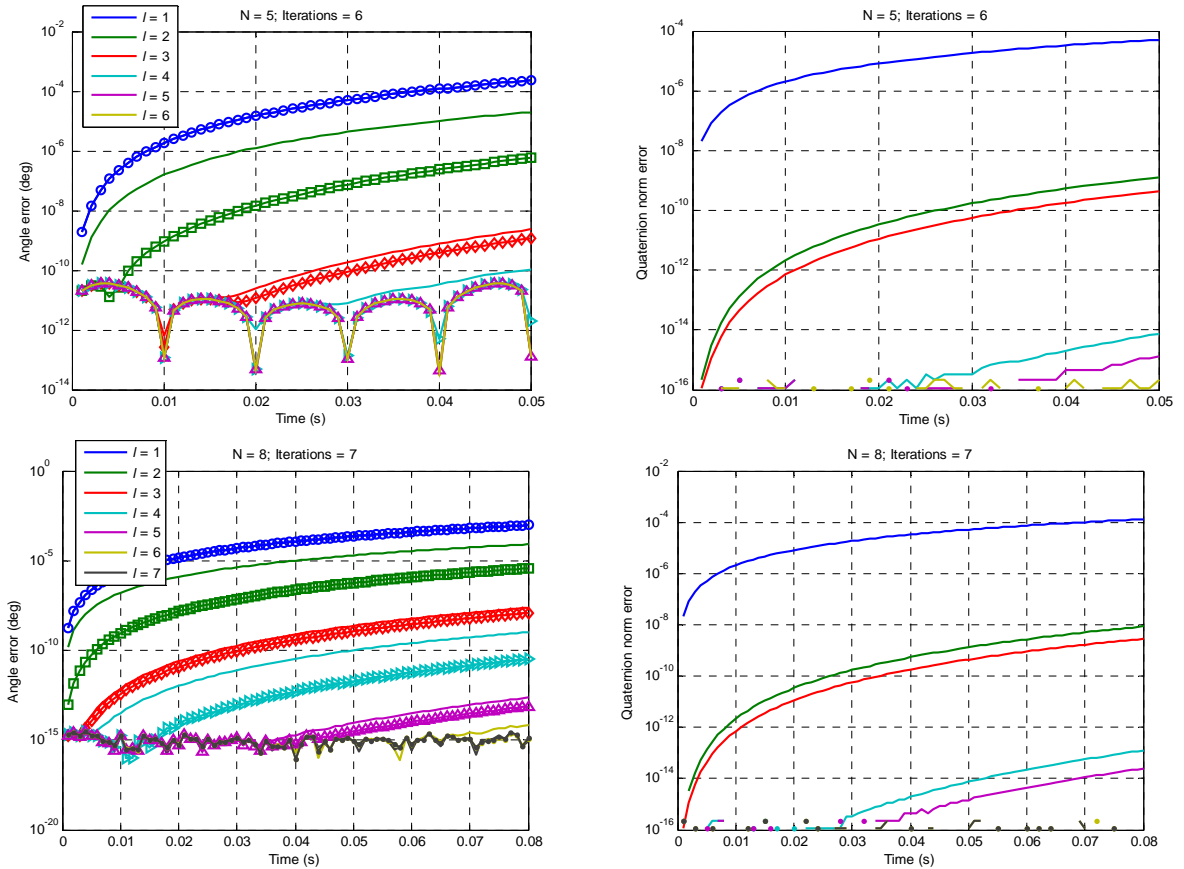

Figure 1. Attitude error comparison of QuatFIter and RodFIter for $N = 5$ (upper row) and 8 (lower row) for coning motion. Decorated lines for RodFIter and simple solid lines for QuatFIter. Right-column subfigures are quaternion estimate norm discrepancy from unity.

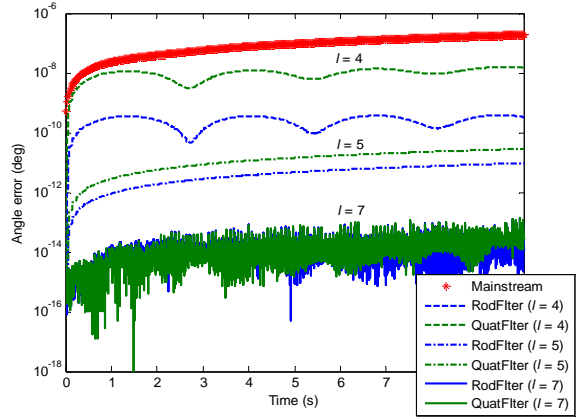

Figure 2. Attitude error comparison of QuatFIter (green lines) and RodFIter (blue lines) for $N = 8$ at 4th, 5th and 7th iterations, as well as the mainstream algorithm ($N = 2$), for coning motion.

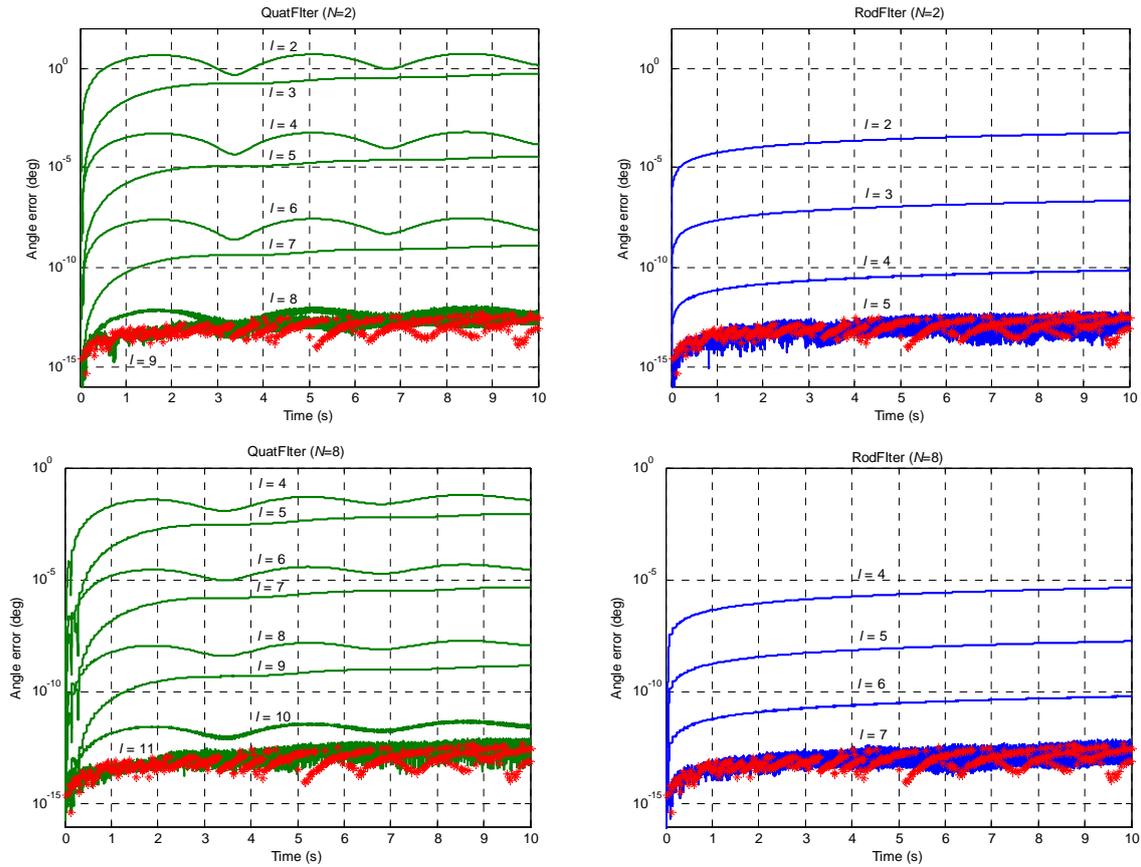

Figure 3. Attitude error comparison of QuatFIter (green lines) and RodFIter (blue lines) for $N = 2$ (upper row) and 8 (lower row), as well as the mainstream algorithm ($N = 2$), for constant angular velocity case $\omega = \begin{bmatrix} 1 & 3 & 2 \end{bmatrix}^T$ rad/s.

Figure 2 presents the attitude computation error of the QuatFIter for ten seconds, compared with the RodFIter ($N$ = 8 at the 4$^{th}$, 5$^{th}$ and 7$^{th}$ iterations) and the mainstream two-sample algorithm in the practical inertial navigation system. It confirms the above observation that the QuatFIter has a slower convergence rate relative to the number of iteration but has final accuracy comparable with the RodFIter. Table II compares the time cost ($N$ = 8 at 7 iterations) averaged across fifty runs on the Matlab platform. It shows that the RodFIter's computational cost is about 2.5 times of that the QuatFIter.

For the special case of constant angular velocity, we set $\boldsymbol{\omega} = \begin{bmatrix} 1 & 3 & 2 \end{bmatrix}^T$ rad/s, for which the non-commutativity vanishes and the mainstream two-sample algorithm is exact. This case belongs to the situation of $\delta\boldsymbol{\omega} = 0$ discussed in the text below (28), where higher truncation order leads to better accuracy, so the truncation order is specially increased to $n_T = n+10$. Figure 3 plots the attitude error of the QuatFIter, as compared with the RodFIter ($N$ = 2 and 8) and the mainstream two-sample algorithm (exact in this special case). Similar phenomenon is observed that the QuatFIter's errors stay consistently above those of RodFIter for the same number of iterations and it takes the QuatFIter four more iterations than the RodFIter to achieve the accuracy level up to machine precision, namely, 9 and 11 iterations versus 5 and 7 iterations for $N$ = 2 and $N$ = 8, respectively. That is to say, the RodFIter has a faster convergence rate and a rather uniform error reduction with respective to the number of iterations. However, the QuatFIter's disadvantage of slower convergence is not a big concern and can be well remedied by spending several more iterations thanks to its faster speed of implementation. Table II also presents the time cost for both to achieve comparable accuracy and the QuatFIter's computational cost further reduces to below one third of the RodFIter's, mainly due to the computational

Table II. Comparison of Computation Time (10-s data, $N$=8)

|  | QuatFIter | RodFIter | Mainstream Algorithm ($N$=2) |
|---|---|---|---|
| Coning motion ($n_T = n+2$) | 1.04s ($l$ = 7) | 2.65s ($l$ = 7) | 0.07s |
| Constant angular velocity ($n_T = n+10$) | 1.47s ($l$ = 11) | 4.83s ($l$ = 7) | |

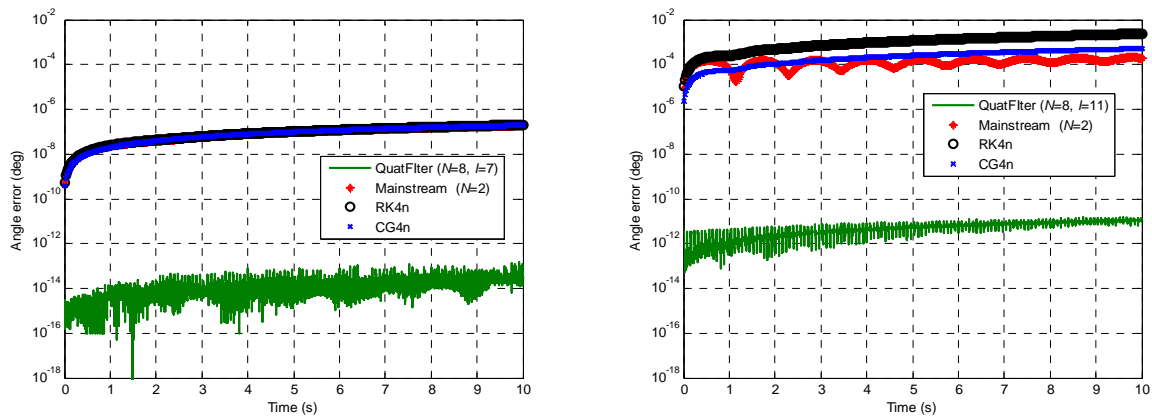

Figure 4. Attitude error comparison of QuatFIter (green lines), the mainstream algorithm ($N = 2$) and two numerical integration methods (RK4n and CG4n), in two cases of $\alpha = 10\deg$, $\Omega = 0.74\pi$ rad/s (left subfigure) and $\alpha = 90\deg$, $\Omega = 1.74\pi$ rad/s (right subfigure).

complexity of the latter proportional to the truncation order squared.

Finally, we compare the QuatFIter with the mainstream two-sample attitude algorithm, as well as two common numerical integration methods addressed in [10], namely, the classical fourth-order Runge-Kutta algorithm (RK4n) and the fourth-order Crouch–Grossman Lie group method (CG4n). The update intervals of RK4n and CG4n are set to be consistent with the mainstream two-sample algorithm. Figures 4 plots the comparison results for two different settings of coning motions: $\alpha = 10\deg$, $\Omega = 0.74\pi$ rad/s and $\alpha = 90\deg$, $\Omega = 1.74\pi$. It shows the RK4n and the CG4n is inferior in accuracy to the mainstream two-sample algorithm, especially in the case of significant coning motions (Fig. 4, right subfigure). This observation accords with the common belief in the navigation community. The QuatFIter performs still the best with a remarkable advantage of at least 6-7 orders in accuracy.

## V. Discussions and Conclusions

There has been a long-time consistent pursuit of accurate attitude computation by integrating angular velocity in many engineering fields. A significant advance has recently appeared in the framework of functional iterative integration of attitude kinematics, for example, the RodFIter based on the Rodrigues vector that is capable of analytically reconstructing the attitude over the whole update time interval. This paper further exploits its benefit and comes up with the QuatFIter for attitude reconstruction using the quaternion in view of its linear kinematic equation. It is shown that the QuatFIter is equivalent to the previous Picard-type successive approximate quaternion method. Chebyshev polynomial approximation and truncation are applied to reduce the computational cost of QuatFIter. Numerical results show that the QuatFIter has about two times better computational efficiency at comparable accuracy to the RodFIter, although the latter has the advantage of relatively uniform and faster error reduction with respect to the number of iterations. It is suspected that the unity-norm constraint of the quaternion contributes to QuatFIter's non-uniform error reduction, but no apparent relationship between the attitude error and the quaternion norm error has been identified so far.

## ACKNOWLEDGEMENTS

Thanks to Dr. Qi Cai for Matlab code optimization.

## APPENDIX

(*Weierstrass M-Test* [22]): Suppose $\{f_j\}_{j=0}^{\infty}$ is a sequence of real functions and the sequence of positive real numbers $\sum_{j=0}^{\infty} \alpha_j$ converges. If $\|f_j(t)\| \leq \alpha_j \, (j \geq 0)$ on a time interval of interest, then the series of real functions $\sum_{j=0}^{\infty} f_j(t)$ converge uniformly and absolutely on the time interval.

*Proposition 1 [24, 25]*: If the commutativity $\mathbf{W}(t_1)\mathbf{W}(t_2) = \mathbf{W}(t_2)\mathbf{W}(t_1)$ holds for any $t_1, t_2 \in [0 \ t]$, then $\mathbf{q}_{\infty} = \exp\left\{\int_0^t \mathbf{W} d\tau\right\} \mathbf{q}(0)$ is the quaternion solution to (4).

Proof. Denote $I_l(t) \triangleq \int_0^t \mathbf{W}(\tau_1) \int_0^{\tau_1} \mathbf{W}(\tau_2) \cdots \int_0^{\tau_{l-1}} \mathbf{W}(\tau_l) d\tau_l \cdots d\tau_2 \, d\tau_1$. It will be proved that $I_l(t) = \frac{1}{l!}\left(\int_0^t \mathbf{W}(\tau) d\tau\right)^l$. This is obviously true for $l = 0, 1$. Assume it is true for $l$, then we have

$$I_{l+1}(t) = \int_0^t \mathbf{W}(\tau) I_l(\tau) d\tau = \int_0^t \mathbf{W}(\tau) \frac{1}{l!}\left(\int_0^t \mathbf{W}(\tau) d\tau\right)^l d\tau \tag{30}$$

The integrand on the right side can be rewritten as

$$\mathbf{W}(\tau) \frac{1}{l!}\left(\int_0^t \mathbf{W}(\tau) d\tau\right)^l = d\left[\frac{1}{(l+1)!}\left(\int_0^t \mathbf{W}(\tau) d\tau\right)^{l+1}\right] \tag{31}$$

which uses the derivative chain rule and the assumed commutativity condition. Therefore $I_{l+1}(t) = \frac{1}{(l+1)!}\left(\int_0^t \mathbf{W}(\tau) d\tau\right)^{l+1}$. Then the limit of the quaternion function sequence of (7) is

$$\begin{aligned}
\mathbf{q}_\infty &= \lim_{l \to \infty} \int_0^t \left( \int_0^{\tau_1} \cdots \left( \int_0^{\tau_{l-1}} \mathbf{q}(0) \circ \frac{\boldsymbol{\omega}}{2} d\tau_l \right) \cdots \circ \frac{\boldsymbol{\omega}}{2} d\tau_2 \right) \circ \frac{\boldsymbol{\omega}}{2} d\tau_1 \\
&= \lim_{l \to \infty} \int_0^t \mathbf{W}(\tau_1) \int_0^{\tau_1} \mathbf{W}(\tau_2) \cdots \int_0^{\tau_{l-1}} \mathbf{W}(\tau_l) d\tau_l \cdots d\tau_2 \, d\tau_1 \cdot \mathbf{q}(0) \\
&= \sum_{l=0}^\infty \frac{1}{l!} \left( \int_0^t \mathbf{W}(\tau) d\tau \right)^l \cdot \mathbf{q}(0) \\
&= \exp\left\{ \int_0^t \mathbf{W} \, d\tau \right\} \mathbf{q}(0)
\end{aligned} \quad (32)$$

It can be readily checked that the above $\mathbf{q}_\infty$ satisfies the quaternion kinematic equation (4).

∎

REFERENCES

[1] P. D. Groves, *Principles of GNSS, Inertial, and Multisensor Integrated Navigation Systems*, 2nd ed.: Artech House, Boston and London, 2013.
[2] D. H. Titterton and J. L. Weston, *Strapdown Inertial Navigation Technology*, 2nd ed.: the Institute of Electrical Engineers, London, United Kingdom, 2007.
[3] F. L. Markley and J. L. Crassidis, *Fundamentals of Spacecraft Attitude Determination and Control*: Springer, 2014.
[4] J. W. Jordan, "An accurate strapdown direction cosine algorithm," NASA TN-D-5384, 1969.
[5] J. E. Bortz, "A new mathematical formulation for strapdown inertial navigation," *IEEE Transactions on Aerospace and Electronic Systems,* vol. 7, pp. 61-66, 1971.
[6] P. G. Savage, *Strapdown Analytics*, 2nd ed.: Strapdown Analysis, 2007.
[7] M. Wang, W. Wu, J. Wang, and X. Pan, "High-order attitude compensation in coning and rotation coexisting environment," *IEEE Trans. on Aerospace and Electronic Systems,* vol. 51, pp. 1178-1190, 2015.
[8] C. Rucker, "Integrating Rotations Using Nonunit Quaternions," *IEEE Robotics and Automation Letters,* vol. 3, pp. 2779-2986, 2018.
[9] J. Park and W.-K. Chung, "Geometric integration on euclidean group with application to articulated multibody systems," *IEEE Trans. on Robotics,* vol. 21, pp. 850-863, 2005.
[10] M. S. Andrle and J. L. Crassidis, "Geometric Integration of Quaternions," *Journal of Guidance, Control, and Dynamics,* vol. 36, pp. 1762-1767, 2013.
[11] M. Boyle, "The Integration of Angular Velocity," *Advances in Applied Clifford Algebras,* vol. 27, pp. 2345–2374, 2017.
[12] P. Krysl and L. Endres, "Explicit Newmark/Verlet algorithm for time integration of the rotational dynamics of rigid bodies," *International Journal for Numerical Methods in Engineering,* vol. 62, pp. 2154–2177, 2005.
[13] E. Hairer, C. Lubich, and G. Wanner, *Geometric Numerical Integration: Structure Preserving Algorithms forOrdinaryDifferential Equations*. New York, NY, USA: Springer-Verlag, 2006.
[14] H. Munthe-Kaas, "Runge-Kutta methods on Lie groups," *BIT,* vol. 38, pp. 92–111, 1998.
[15] P. Savage, "Down-Summing Rotation Vectors For Strapdown Attitude Updating (SAI WBN-14019)," Strapdown Associates (http://strapdownassociates.com/Rotation%20Vector%20Down_Summing.pdf) 2017.7.
[16] Y. Wu, "RodFIter: Attitude Reconstruction from Inertial Measurement by Functional Iteration," *IEEE Trans. on Aerospace and Electronic Systems,* vol. 54, pp. 2131-2142, 2018.
[17] Y. Wu, Q. Cai, and T.-K. Truong, "Fast RodFIter for Attitude Reconstruction from Inertial Measurement," *to appear in IEEE Trans. on Aerospace and Electronic Systems (early access: https://ieeexplore.ieee.org/document/8438980),* 2019.2.
[18] G. Yan, J. Weng, X. Yang, and Y. Qin, "An Accurate Numerical Solution for Strapdown Attitude Algorithm based on Picard iteration," *Journal of Astronautics,* vol. 38, pp. 65-71, 2017.
[19] Z. Xu, J. Xie, Z. Zhou, J. Zhao, and Z. Xu, "Accurate Direct Strapdown Direction Cosine Algorithm," *to appear in IEEE Trans. on Aerospace and Electronic Systems (early access: https://ieeexplore.ieee.org/document/8534467),* 2018.
[20] V. N. Branets and I. P. Shmyglevsky, *Introduction to the Theory of Strapdown Inertial Navigation System*: Moscow, Nauka (in Russian), 1992.
[21] J. Kuipers, *Quaternions and Rotation Sequences*: Princeton University Press, NJ, 1998.
[22] W. J. Rugh, *Linear System Theory*, 2nd ed. New Jersey: Prentice-Hall, 1996.
[23] W. H. Press, *Numerical Recipes: the Art of Scientific Computing*, 3rd ed. Cambridge ; New York: Cambridge University Press, 2007.
[24] R. W. Brockett, *Finite Dimensional Linear Systems*. New York: Wiley, 1970.
[25] M. Baake and U. Schlaegel, "The Peano Baker Series," *Proceedings of the Steklov Institute of Mathematics,* vol. 275, pp. 155–159, 2011.